%% file: forcewam.tex
\documentclass{article}

\usepackage{forcewam_preprint,times}
\input{math_commands.tex}

\usepackage{amsmath}
\usepackage{amssymb}
\usepackage{booktabs}
\usepackage{float}
\usepackage{graphicx}
\usepackage{capt-of}
\usepackage{wrapfig}
\usepackage{needspace}
\usepackage{placeins}
\usepackage{longtable}
\usepackage{multirow}
\usepackage{array}
\usepackage{enumitem}
\usepackage[nopatch=eqnum]{microtype}
\usepackage{xcolor}
\usepackage{hyperref}
\usepackage{url}
\hypersetup{
    colorlinks=true,
    allcolors=black,
    pdftitle={ForceWAM: Decoupling Future and Response Learning from Demo and Play Data for Contact-Rich Manipulation},
    pdfauthor={Zijun Xu, Wendi Chen, Xinkai Wang, Jiachen Zou, Hongjin Chen, Hairuo Liu, Bokai Lin, Shihao Ma, Ke Ma, Zhiwei Zhang, Zhuohao Yan, Ke Wu, Haoming Song, Xueqi Li, Chuan Wen, Jieru Zhao, Wenchao Ding, Cewu Lu, Lixin Yang}
}
\title{ForceWAM: Decoupling Future and Response\\
Learning from Demo and Play Data for\\
Contact-Rich Manipulation}

\author{%
\begin{minipage}{\dimexpr\textwidth-2\tabcolsep\relax}
\centering\normalfont\normalsize
Zijun Xu$^{1,4,*}$\quad \href{https://wendichen.me/}{Wendi Chen}$^{2,4,*}$\quad \href{https://summerwxk.github.io/}{Xinkai Wang}$^{3,4,*}$\quad Jiachen Zou$^{2}$\\
Hongjin Chen$^{1}$\quad Hairuo Liu$^{2,4}$\quad Bokai Lin$^{2,4}$\quad Shihao Ma$^{1}$\quad Ke Ma$^{1}$\\
Zhiwei Zhang$^{1}$\quad \href{https://www.zhhyan.cn/}{Zhuohao Yan}$^{1}$\quad Ke Wu$^{1}$\quad Haoming Song$^{2}$\quad Xueqi Li$^{4,5}$\\
\href{https://alvinwen428.github.io/}{Chuan Wen}$^{2}$\quad \href{https://zjru.github.io/}{Jieru Zhao}$^{2}$\quad \href{https://wenchaoding.github.io/personal/index.html}{Wenchao Ding}$^{1}$\quad \href{https://www.mvig.org/}{Cewu Lu}$^{2,4,6}$\quad \href{https://lixiny.github.io/}{Lixin Yang}$^{2,\dagger}$\\[0.6em]
\small
$^{1}$\href{https://fudan.edu.cn}{Fudan University}\quad
$^{2}$\href{https://sjtu.edu.cn}{Shanghai Jiao Tong University}\\
$^{3}$\href{https://seu.edu.cn}{Southeast University}\quad
$^{4}$\href{https://sii.edu.cn}{Shanghai Innovation Institute}\\
$^{5}$\href{https://sustech.edu.cn}{Southern University of Science and Technology}\quad
$^{6}$Noematrix Ltd.\\[0.4em]
\footnotesize $^{*}$Equal contribution.\quad $^{\dagger}$Corresponding author.
\end{minipage}%
}

\newcommand{\method}{ForceWAM}
\newcommand{\projecturl}{%
  {\hypersetup{urlcolor=blue}\urlstyle{same}%
   \textbf{\url{https://force-wam.github.io/}}}%
}

\begin{document}

\raggedbottom

\maketitle
\pagestyle{plain}

\vspace{-1em}
{\centering
\small\textbf{Project page:} \projecturl\par
}
\vspace{1em}

\begin{figure}[H]
    \centering
    \includegraphics[width=\linewidth]{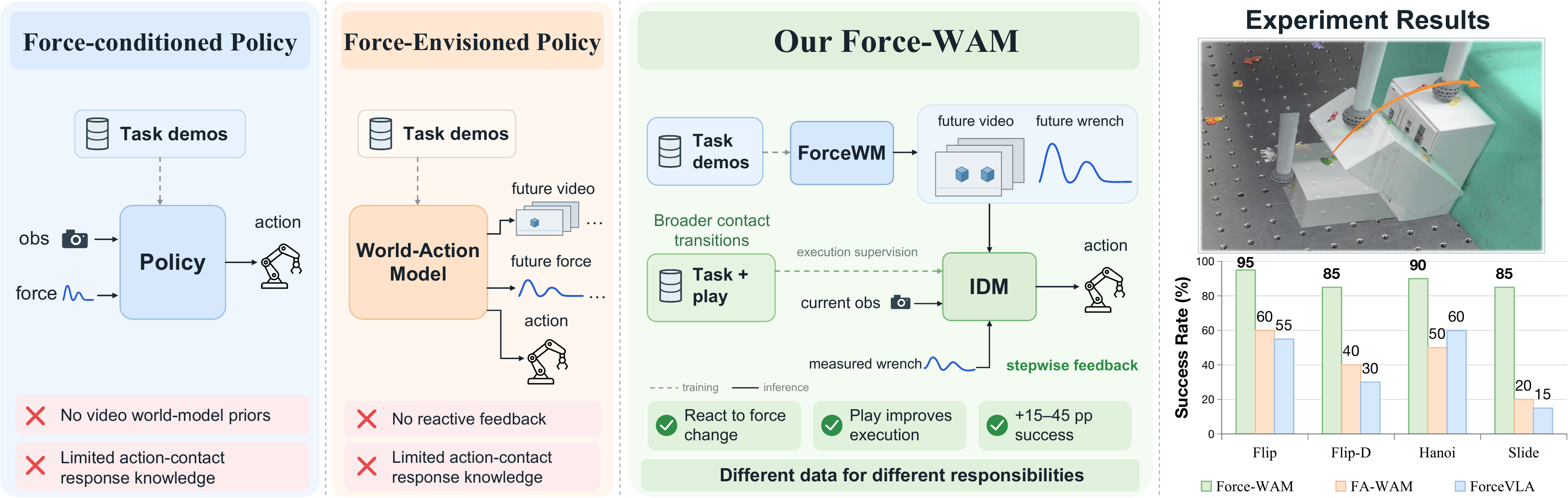}
    \caption{\textbf{Motivation of ForceWAM.}
    The illustrated end-to-end force-conditioned and force-envisioned policies use only task demonstration data:
    the former lack pretrained video-world-model priors, while the latter lack
    within-chunk force feedback. ForceWAM combines the pretrained prior with task
    demonstrations for task-compatible visual-wrench futures and action-labeled
    play for IDM execution learning, achieving the highest mean success across
    all four real-robot conditions.}
    \label{fig:forcewam-teaser}
\end{figure}

\begin{abstract}
\looseness=-1
Expert demonstrations show how to complete contact-rich tasks, but provide
limited coverage of how interaction forces respond to different actions.
Task-agnostic play provides broader experience of contact dynamics without
specifying task goals. To fully leverage these two types of data, we introduce
\method{}, a force world action model that decouples knowledge of
task-compatible interaction outcomes from knowledge of how local actions change
contact. A world model learns from successful demonstrations to jointly predict future video
and six-axis force/torque trajectories. Together, these futures
describe the expected evolution of motion and contact during the task. A separate
inverse dynamics model (IDM) learns from both task and play,
mapping current observations and recorded visual-wrench futures to their
associated robot actions. Play data thus provide knowledge of force responses
for goal-directed execution. At deployment, the IDM reactively translates
generated futures into actions, reusing them with fresh wrench feedback between
world-model updates. Across real-robot force-regulated sliding, box flipping
with and without disturbances, and ring placement, \method{} achieves
$85$-$95\%$ mean success, surpassing world-modeling and reactive baselines.
Our experiments
further show that adding task-agnostic play data improves success by
$15$-$45$ percentage points, demonstrating its value for learning
contact-rich execution.
Project page and code are available at \projecturl.
\end{abstract}

\input{sections/01_introduction}
\input{sections/02_related_work}

\input{sections/03_method}
\begingroup
\setlength{\textfloatsep}{6pt plus 2pt minus 2pt}
\input{sections/04_experiments}

\Needspace{5\baselineskip}
\input{sections/05_conclusion}

\clearpage
\endgroup
\bibliography{forcewam}
\bibliographystyle{forcewam}

\clearpage
\input{sections/appendix}

\end{document}

%% file: math_commands.tex
\usepackage{amsmath,amsfonts,bm}

\def\eqref#1{equation~\ref{#1}}

\def\1{\bm{1}}

\DeclareMathAlphabet{\mathsfit}{\encodingdefault}{\sfdefault}{m}{sl}
\SetMathAlphabet{\mathsfit}{bold}{\encodingdefault}{\sfdefault}{bx}{n}

%% file: sections/01_introduction.tex
\section{Introduction}
\label{sec:introduction}

Robotic manipulation is often cast as predicting actions that reach a desired
visual state. Yet contact-rich tasks are decided not only by where the robot
moves, but by how interaction forces evolve along the way. Two visually similar
configurations can correspond to insufficient contact, damaging force, or a
jam. Classical force control therefore places interaction feedback at the
center of execution \citep{hogan1985impedance}, while recent learned systems
show that force and touch improve contact-rich manipulation
\citep{liu2025forcemimic,xue2025rdp,yu2025forcevla,chen2026implicitrdp,wang2026lift,zhuo2026fardp}.
We call this learning problem \emph{force-centric manipulation}: a policy must
not merely sense force, but learn the physical regularities needed to regulate
it. Recent methods learn adaptive compliance or hybrid force-position control
\citep{hou2024acp,fang2026forcepolicy,li2026forcevla2}. We aim to learn the
full force-centric manipulation task directly from data. However, the main obstacle is a mismatch
between where these regularities appear in data and how policies learn them.

\looseness=-1
Expert demonstrations, which we call \emph{task data}, show successful
task execution. A box-pushing demonstration at roughly $10\,$N, for example,
need not show how force changes when the robot presses farther, backs away,
or encounters a disturbance. By contrast, \emph{task-agnostic play data}
record exploratory pressing, sliding, and release, providing additional
visual-wrench transitions and associated robot actions.
Play transitions need not demonstrate successful behavior for a given task
instruction, but they still provide recorded future-action pairs for
inverse-dynamics learning. End-to-end force-aware learners
\citep{yu2025forcevla,chen2026implicitrdp} still couple two questions in one
task-supervised mapping: \emph{what} transition advances the task, and
\emph{which action} realizes that transition. Our question is: can we use these two data sources more effectively to learn the
distinct knowledge each provides?

Video planning with inverse dynamics provides a foundation for learning actions
from interaction outcomes: recorded transitions supervise the inverse model,
while generated futures guide execution \citep{du2023unipi,li2026vera}.
For contact-rich manipulation, a visual future can leave the desired evolution
of contact forces unspecified. Force- and tactile-aware models have demonstrated
the value of predicting contact signals and using them for reactive control
\citep{he2026fawam,ma2026tacpac}. We study visual-wrench futures as the
condition for learning execution from task and play, associating actions with
both the motion and the contact evolution they produce.

We introduce \textbf{\method{}} (Figure~\ref{fig:forcewam-teaser}), which uses
visual-wrench futures to decouple knowledge of task-compatible interaction
outcomes from knowledge of how local actions change contact. Successful task
demonstrations train the force world model (ForceWM) to jointly generate future video and six-axis
wrench trajectories. The inverse
dynamics model (IDM) learns from task and play segments to predict the robot
actions that realize a visual-wrench future. During training, it
conditions on the current observation and the future recorded in each segment.
Any segment that records actions and their resulting motion and contact
outcomes can supervise this action prediction, even if it does not complete a
task.
At deployment, ForceWM-generated futures replace the recorded future conditions
used in IDM training, specifying the interaction that execution should realize.
The IDM reuses each generated reference within an action chunk, reactively
updating its predictions with newly measured wrench feedback.

We evaluate \method{} on force-regulated sliding, box flipping with and
without disturbances, and ring placement, achieving $85$-$95\%$ mean
success across the four evaluation conditions. Our experiments assess
the benefit of added play and the role of visual-wrench future conditioning
in execution. Our contributions are:
\begin{itemize}[topsep=4pt,partopsep=0pt,itemsep=3pt,parsep=0pt]
    \item We formulate contact execution as inverse dynamics conditioned on
    visual-wrench futures. This formulation allows different types of
    interaction data to supervise action prediction.
    \item We propose \method{}, whose visual-wrench future interface decouples
    knowledge of task-compatible interaction outcomes from knowledge of local
    action-contact responses.
    ForceWM learns from successful demonstrations. The IDM learns from task
    and play, mapping generated futures to actions with causal wrench
    feedback between refreshes.
    \item Across four real-robot evaluation conditions, we show that added
    play improves success by $15$-$45$ percentage points with task
    demonstrations, ForceWM, and IDM optimizer updates held fixed.
    Future-conditioning ablations further show the benefit of wrench futures
    alongside visual futures, with measured wrench feedback retained.
\end{itemize}

%% file: sections/02_related_work.tex
\section{Related Work}
\label{sec:related}

\paragraph{Force-aware manipulation.}
\looseness=-1
Force and tactile feedback provide contact information that is difficult to
infer from vision alone \citep{lee2019makingsense,suresh2024neuralfeels,ai2024robopack}.
ForceMimic couples force-motion demonstration capture
with hybrid imitation, while ForceVLA integrates force into
vision-language-action policies and ForceVLA2 extends this line to hybrid
force-position control \citep{liu2025forcemimic,yu2025forcevla,li2026forcevla2}.
\mbox{Beyond incorporating} force as an observation, reactive policies use newly
measured contact signals to update actions during execution \citep{calandra2018feeling}. RDP combines a
slow latent policy with a fast tactile executor, whereas ImplicitRDP realizes
within-chunk force feedback through causal attention in a single diffusion
network \citep{xue2025rdp,chen2026implicitrdp}. FA-RDP adapts inference
frequency to manipulation phase, and LIFT combines reactive force injection
with online corrective supervision \citep{zhuo2026fardp,wang2026lift}.
\method{} decouples knowledge of task-compatible interaction outcomes from
knowledge of how local actions change contact:
ForceWM models what a successful interaction should look and feel like, while
an inverse dynamics model (IDM) turns that future into robot actions and
reactively updates them with within-chunk wrench feedback.
The IDM is conditioned on the current observation and visual-wrench future,
which lets it learn from both task and play transitions.

\paragraph{World action models.}
World action models connect future observation prediction with robot action
generation \citep{bi2026motus,bi2026motus2,cai2026ahawam}, as exemplified by LingBot-VA, Cosmos Policy, DreamZero, FACT, and
GE-Act 2.0
\citep{li2026lingbotva,kim2026cosmospolicy,ye2026dreamzero,peng2026fact,agibot2026geact2}. A related route
separates video planning from action inference \citep{xu2025vilp}: UniPi translates generated
visual plans with a separately trained IDM, and VERA develops a Jacobian-based
IDM that can learn independently from self-play \citep{du2023unipi,li2026vera}.
Contact-aware models extend prediction to force and tactile dynamics for
planning and reactive control
\citep{higuera2026vtwm,zheng2026omnivta,zhou2026touchworld,madan2026tactic,neoteai2026n0twam}.
Among these, FAWAM jointly predicts actions and wrenches and trains residual
corrections from human interventions and policy rollouts
\citep{he2026fawam}. TacPAC uses live tactile observations and cached
tactile-action predictions to correct an existing action chunk, with its
corrector supervised against demonstrated actions \citep{ma2026tacpac}.
\method{} separates learning task-compatible outcomes from learning local
action-contact responses while retaining a force-aware reactive inference process.
Successful demonstrations train
a visual-wrench future generator, while task and play windows directly train an
IDM on their recorded futures and associated actions. At deployment, the IDM
uses the predicted visual-wrench future as a task-conditioned reference and
reactively corrects each action with newly measured wrench feedback within the
cached future window. Thus, the two modules are decoupled in what they learn but
connected through force feedback during execution, rather than using a visual
future only as a static planning target. This makes task-agnostic contact
transitions a direct source of supervision for executing task-conditioned
visual-wrench futures.

\suppressfloats[t]

%% file: sections/03_method.tex
\section{Method}
\label{sec:method}

\subsection{Overview}

\method{} separates two questions that are entangled in an end-to-end policy:
\emph{what should a successful interaction look and feel like?}, and
\emph{what robot actions can realize that interaction?}  Let
$\mathbf o_t=(\mathbf I_t,\mathbf w_t)$ denote the current visual-wrench
observation, let $\mathbf g_t$ denote a future visual-wrench reference, and
let $\mathbf A_t=(\mathbf a_{t,0},\ldots,\mathbf a_{t,H_a-1})$ denote a short
action sequence.  Our architecture factorizes their joint distribution as
\begin{align}
    p(\mathbf g_t,\mathbf A_t\mid \mathbf o_t,\ell)
      &=p(\mathbf g_t\mid \mathbf o_t,\ell)\,
        p(\mathbf A_t\mid \mathbf o_t,\mathbf g_t,\ell)
        \notag\\
      &\approx p_\phi(\mathbf g_t\mid \mathbf o_t,\ell)\,
        p_\theta(\mathbf A_t\mid \mathbf o_t,\mathbf g_t),
        \label{eq:wam-idm-factorization}
\end{align}
where the first factor is ForceWM and the second is the inverse dynamics model
(IDM).  The IDM predicts actions from the current observation and a
visual-wrench future, without a separate task instruction $\ell$.  At
deployment, ForceWM supplies the task-conditioned future $\mathbf g_t$, and
the IDM reactively updates actions using newly measured wrench feedback.
This conditioning lets task and play data share the same inverse-dynamics
objective.  We train $p_\phi$ on successful task trajectories
$\mathcal D_{\mathrm{task}}$, whereas $p_\theta$ can learn from
$\mathcal D_{\mathrm{task}}\cup\mathcal D_{\mathrm{play}}$.

\begin{figure}[t]
    \centering
    \includegraphics[width=\linewidth,trim=26bp 44bp 44bp 38bp,clip]{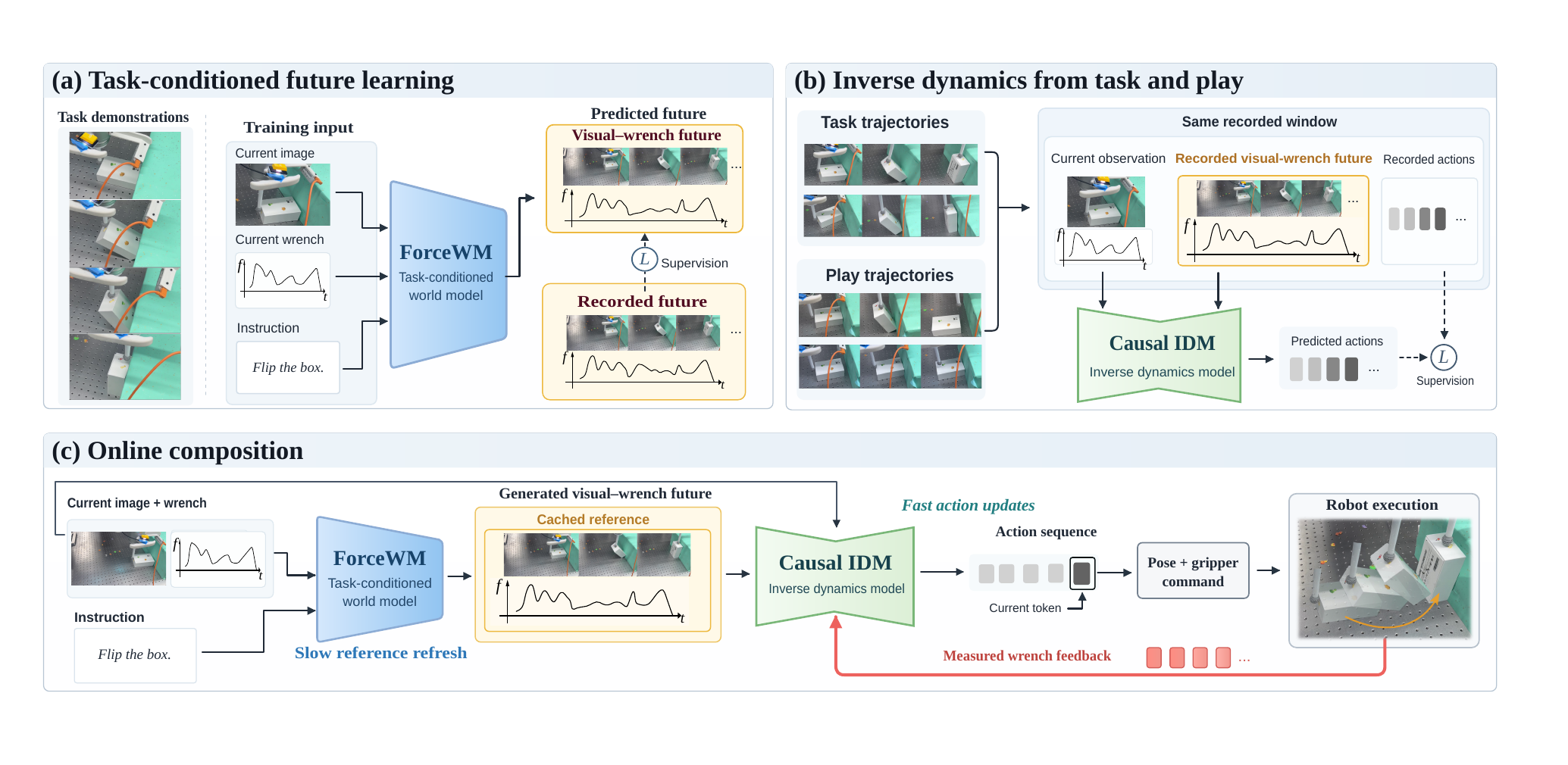}
    \caption{\textbf{\method{} training and deployment.} Task demonstrations train the
    visual-wrench future generator, while task and play data train the IDM
    from recorded futures and associated actions. During deployment, ForceWM
    supplies future conditions and the causal IDM updates actions using
    measured wrench feedback.}
    \label{fig:forcewam-architecture-overview}
    \vspace{-8pt}
\end{figure}

At time $t$, let $\mathbf I_t$ denote the wrist image,
$\mathbf w_t\in\mathbb R^6$ the measured TCP force and torque, and $\ell$ the
task instruction.  We use two trajectory sources.  Successful \emph{task data}
show futures compatible with the task.  \emph{Play data} contain the same
synchronized image, wrench, TCP-pose, and gripper-state streams, but need not
solve or even attempt the downstream task.  We therefore use play to learn
local execution responses.

\subsection{Prediction from task data}

For a trajectory window starting at time $t$, we construct the future
prediction target
$\mathbf g_t=(\mathbf z_t,\mathbf W_t)$,
$\mathbf W_t=(\bar{\mathbf w}_{t+1},\ldots,
              \bar{\mathbf w}_{t+H_g})$,
where $\mathbf z_t$ encodes future wrist images and $\mathbf W_t$ is a
normalized future wrench trajectory.  Thus, $\mathbf g_t$ is the target visual
and contact trajectory for ForceWM. The corresponding robot actions are
modeled separately by the IDM. Appendix~\ref{app:representation} specifies
frame offsets and encodings.

ForceWM models
$p_\phi(\mathbf g_t\mid\mathbf I_t,\mathbf w_t,\ell)$
from successful task demonstrations.  Its backbone is a pretrained
spatiotemporal video diffusion model.  We arrange its latent input as five
semantic slots: a blank prefix, the current wrench, the current image, the
future wrench trajectory, and the future visual latent.  The first three slots
are clamped throughout denoising, while the last two are jointly generated.
A fixed packing adapter repeats low-dimensional wrench values over the spatial
extent of a video-latent slot.  This construction reuses one spatiotemporal backbone for the two
future modalities while retaining separate visual and wrench outputs.

For a clean task future $\mathbf g_t$, we sample a noise level $\sigma$ and
train an Elucidated Diffusion Model (EDM) denoiser
\citep{karras2022elucidating} to recover the joint future,
\begin{align}
    \mathcal L_{\mathrm{WM}}
      &=\mathbb E\!\left[
        \lambda(\sigma)
        \left\|\mathbf M_{\mathrm{gen}}\odot\left(
        \mathbf x_t-D_\phi(\mathbf x_t^\sigma;
        \ell,\sigma)\right)\right\|_2^2
      \right].
      \label{eq:forcewm-loss}
\end{align}
Here $\mathbf x_t$ is the five-slot clean latent,
$\mathbf x_t^\sigma$ adds Gaussian noise only to its generated slots, and
$\mathbf M_{\mathrm{gen}}$ selects the future-wrench and future-visual slots.
The image and wrench conditions enter $D_\phi$ through their clamped slots,
whereas $\ell$ enters through text conditioning.  Consequently, the blank,
current-wrench, and current-image slots remain clean and receive no
reconstruction loss.  EDM training samples $\sigma$ from a hybrid log-normal
and high-noise uniform distribution. The exact mixture is given in
Appendix~\ref{app:training-details}. At inference, ForceWM starts
the two future slots from noise and samples
$\widehat{\mathbf g}_t=(\widehat{\mathbf z}_t,
\widehat{\mathbf W}_t)$.  Training on successful demonstrations targets a
distribution of task-compatible interactions rather than a single force
trace. Multiple visual-force futures can accomplish the task.

\subsection{Inverse dynamics learning from task and play}
\label{sec:idm-learning}

The IDM models a conditional distribution over a short sequence
$\mathbf A_t=(\mathbf a_{t,0},\ldots,\mathbf a_{t,H_a-1})$.  The executable
part of each action contains a Cartesian TCP-pose target and a gripper command.
As in ImplicitRDP \citep{chen2026implicitrdp}, training also predicts
force-derived auxiliary targets that are discarded at execution. These targets
are held fixed across all IDM comparisons.  TCP poses are represented relative
to the pose at the beginning of the window. The exact target construction and
deterministic conversion to robot commands are given in
Appendix~\ref{app:action-interface}.

Each IDM training window contains the current image $\mathbf I_t$, the recorded
future $\mathbf g_t$ constructed from later frames and wrenches in that same
window, the action-rate wrench sequence, and the associated action sequence
$\mathbf A_t$.  ForceWM is not run to create IDM training inputs.  The same
construction applies to task and play trajectories: although a play window
need not accomplish the downstream task, it still provides one feasible
action sequence associated with its realized visual-wrench future.

The conditional diffusion IDM \citep{chi2023diffusionpolicy} uses normalized
actions $\bar{\mathbf A}_t$. At step $q$, define
$\mathbf A_t^q
    =\sqrt{\bar\alpha_q}\,\bar{\mathbf A}_t
     +\sqrt{1-\bar\alpha_q}\,\boldsymbol\epsilon$
and
$\mathbf v_q
    =\sqrt{\bar\alpha_q}\,\boldsymbol\epsilon
     -\sqrt{1-\bar\alpha_q}\,\bar{\mathbf A}_t$.
We minimize
\begin{equation}
    \mathcal L_{\mathrm{IDM}}
    =\mathbb E\!\left[
      \sum_{k=0}^{H_a-1}
      \left\|\mathbf v_{q,k}-
      \mathbf v_{\theta,k}\!\left(
        \mathbf A_t^q,q;
        \mathbf I_t,\widetilde{\mathbf g}_t,
        \mathbf w_{t:t+k}
      \right)\right\|_2^2
    \right].
    \label{eq:idm-loss}
\end{equation}
Equation~\ref{eq:idm-loss} is written token-wise to expose the feedback available
for action offset $k$.  The Transformer evaluates all action tokens in parallel using
the causal structure described next.  Exact token dimensions, network depth,
and diffusion schedule appear in Appendix~\ref{app:training-details}.

\paragraph{Future-conditioning augmentation.}
To reduce the IDM's sensitivity to imperfect ForceWM predictions at deployment,
we apply random dropout and perturbations of the recorded visual-wrench future.
We pair the augmented condition $\widetilde{\mathbf g}_t$ with the original
action targets (Appendix~\ref{app:training-details}).

\subsection{Slow-fast execution of \method{}}

ForceWM updates slowly, but the IDM can rapidly adjust actions with new wrench
feedback. Thus, we use a temporally causal structure to enable streaming IDM
inference. Furthermore, the slow-fast executor coordinates periodic ForceWM
predictions with these reactive action updates.

\paragraph{Temporally causal IDM.}
The IDM separates conditions by update rate.  For a training window, slow
encoders map the current observation and regularized recorded future to fixed
tokens, while a recurrent fast encoder represents the wrench prefix available
at action offset $k$,
$\widetilde{\mathbf C}_t^{\mathrm{slow}}
      =E_{\mathrm{slow}}(\mathbf I_t,\mathbf w_t,
                         \widetilde{\mathbf g}_t)$,
$\mathbf C_{t,k}^{\mathrm{fast}}
      =E_{\mathrm{fast}}(\mathbf w_{t:t+k})$.
The current image is encoded by a convolutional visual backbone, the current
wrench by a linear adapter, a separate convolutional adapter encodes the future
visual latent, and linear adapters encode the future wrench samples.
A unidirectional recurrent encoder produces one fast token per measured wrench,
enabling streaming action prediction as new feedback arrives.

Training processes the full synchronized action and wrench window in parallel,
but applies lower-triangular masks to both action self-attention and
action-to-measured-wrench attention. Action token $k$ accesses only noisy action
tokens $0{:}k$ and measured wrench tokens $0{:}k$, ensuring that its inputs are
available during streaming inference.
A complementary mask hides reference-wrench tokens at window offsets
$\leq k$, leaving only later reference samples visible to that action token.
This reduces the training-to-inference shift when generated references replace
recorded futures.  Future visual tokens remain globally visible.  These masks
implement the token-wise condition in Equation~\ref{eq:idm-loss} without
future-feedback leakage.

\paragraph{ForceWM-IDM slow-fast execution.}
After training, the recorded future is replaced by a ForceWM prediction.  In
the synchronous inference process, let $r$ denote the sensor request that starts
a slow update and $s$ the control index at which its first new command becomes
active.  ForceWM generates $\widehat{\mathbf g}_r$ from the request-time image,
wrench, and task instruction.  Because its representation matches the recorded
future used for IDM training, it passes through the same slow encoders.
Besides, we sample the IDM's initial diffusion-noise tensor,
\begin{equation}
    \widehat{\mathbf g}_r
      \sim p_\phi(\mathbf g_r\mid\mathbf I_r,\mathbf w_r,\ell),
    \quad
    \widehat{\mathbf C}^{\mathrm{slow}}_r
      =E_{\mathrm{slow}}(\mathbf I_r,\mathbf w_r,
                         \widehat{\mathbf g}_r),
    \quad
    \boldsymbol\xi_r\sim\mathcal N(\mathbf0,\mathbf I).
    \label{eq:slow-cache}
\end{equation}
The generated reference, slow tokens, diffusion noise, and base TCP pose
$\mathbf T_r$ are cached until the next ForceWM update. At action offset $k$, reference-wrench
visibility follows the same window-indexed mask.  With the current feedback
prefix $\mathbf h_{r,s}^{(k)}=(\mathbf w_r,\mathbf w_{s+1},\ldots,
\mathbf w_{s+k})$, the IDM deterministically re-denoises from the same cached
noise:
\begin{equation}
    \widehat{\mathbf a}_{s,k}
      =\bigl[F_\theta\!\bigl(
          \boldsymbol\xi_r;
          \widehat{\mathbf C}^{\mathrm{slow}}_r,
          E_{\mathrm{fast}}(\mathbf h_{r,s}^{(k)})\bigr)\bigr]_k,
    \qquad
    \mathbf u_{s+k}
      =\operatorname{Decode}_{\mathrm{exec}}
        (\widehat{\mathbf a}_{s,k};\mathbf T_r).
    \label{eq:fast-execution}
\end{equation}
Only the current action token is decoded and sent to the low-level controller.
The cached base TCP pose $\mathbf T_r$ converts the relative Cartesian prediction
to an absolute pose-gripper command. Within an action chunk, reusing
$\boldsymbol\xi_r$ with deterministic DDIM sampling removes variation caused
solely by resampling, while the growing measured-wrench feedback prefix changes
the conditioning of the newly predicted action token.  Deployment settings and inference times are reported
in Appendix~\ref{app:deployment-details}.

%% file: sections/04_experiments.tex
\section{Experiments}
\label{sec:experiments}

We organize the evaluation around four questions.  (Q1) Does the complete
system improve real-robot contact-rich manipulation?  (Q2) How does IDM
execution change as it gains access to more task-agnostic play?  (Q3) What
visual and contact evolution does ForceWM predict during closed-loop execution?
And (Q4) how do visual and wrench future conditions contribute to execution?

\input{sections/figs/real_robot_comparisons}

\subsection{Experimental setup}
\label{sec:experiment-setup}

\paragraph{Tasks.}
We evaluate three real-robot task families under four contact-rich conditions.  \emph{Sliding} drags the end
effector along the centerline of a force scale while
targeting a normal force of $10\,$N. Success requires full-path completion and
satisfaction of the $5$-$15\,$N contact-force criterion.
\emph{Box flipping} requires sustained contact and a controlled pushing force
to carry the box over its edge, and its disturbed variant (Flip-D) adds an
external push during contact.
\emph{Hanoi ring placement} seats a held ring onto a $10\,$mm pole, where
contact force decides whether the ring seats or jams.  All four conditions are
contact-rich and use six-axis wrench feedback. They
differ in the kind of contact they demand---continuous force maintenance,
sustained pushing, rejection of an external disturbance, and fine seating contact.

\paragraph{Datasets.}
We collect $200$ task demonstrations and $200$ play episodes for each of the
three task families. Flip and Flip-D share the same training data.
Both sources provide synchronized wrist RGB, six-axis TCP wrench measurements,
and robot actions.  Task demonstrations
record successful task executions.  Play is collected without an
evaluation-task instruction and consists of exploratory pressing, sliding,
and making and breaking contact.  These recorded visual-wrench transitions
provide additional supervision for IDM training.

\paragraph{Baselines.}
We compare against force-reactive policies RDP \citep{xue2025rdp} and
ImplicitRDP \citep{chen2026implicitrdp}, along with ForceVLA
\citep{yu2025forcevla}. We also include world-action models Cosmos Policy
\citep{kim2026cosmospolicy}, Fast-WAM \citep{yuan2026fastwam}, and FA-WAM
\citep{he2026fawam}, as well as $\pi_{0.5}$
\citep{physicalintelligence2025pi05}. In
Table~\ref{tab:task-outcomes}, unmarked rows use task demonstrations without
action-labeled play, and $+$Play marks the variants additionally given play.
For the $+$Play variants of ImplicitRDP and Cosmos Policy, task demonstrations
are augmented with play trajectories.  Cosmos Policy uses an empty language
condition for play and learns to predict the recorded actions.

\paragraph{Evaluation.}
Each evaluation run consists of $20$ trials per task from matched
initial states, under the same safety limits and controller interface.  A trial
succeeds if the task completes without a safety stop or human intervention. For
Sliding, success additionally requires full-path completion and satisfying the
$5$-$15\,$N contact-force criterion. We report mean success rates.

\subsection{Q1: Overall real-robot performance}
\label{sec:overall-performance}

We compare \method{} with the baselines across the four real-robot conditions
to assess task completion under different contact demands.
Table~\ref{tab:task-outcomes} summarizes success rates.

\begin{table}[!htb]
\vspace{-12pt}
\caption{\textbf{Mean real-robot success rates} (the best per column in bold).
Slide requires full-path completion with $5$-$15\,$N contact force.
``$+$Play'': training with additional action-labeled play data.}
\label{tab:task-outcomes}
\centering
\footnotesize
\setlength{\tabcolsep}{4.5pt}
\renewcommand{\arraystretch}{1}
\begin{tabular*}{\linewidth}{@{\extracolsep{\fill}}lcccc@{}}
\toprule
\textbf{Method} & \textbf{Slide}$\,\uparrow$ & \textbf{Flip}$\,\uparrow$ & \textbf{Flip-D}$\,\uparrow$ & \textbf{Hanoi}$\,\uparrow$ \\
\midrule
RDP~\citeyearpar{xue2025rdp} & 15\% & 40\% & 30\% & 45\% \\
ImplicitRDP~\citeyearpar{chen2026implicitrdp} & 20\% & 45\% & 25\% & 55\% \\
ImplicitRDP~\citeyearpar{chen2026implicitrdp} $+$ Play & 10\% & 25\% & 10\% & 30\% \\
ForceVLA~\citeyearpar{yu2025forcevla} & 15\% & 55\% & 30\% & 60\% \\
Cosmos Policy~\citeyearpar{kim2026cosmospolicy} & 15\% & 40\% & 30\% & 60\% \\
Cosmos Policy~\citeyearpar{kim2026cosmospolicy} $+$ Play & 10\% & 20\% & 15\% & 40\% \\
Fast-WAM~\citeyearpar{yuan2026fastwam} & 10\% & 50\% & 25\% & 65\% \\
FA-WAM~\citeyearpar{he2026fawam} & 20\% & 60\% & 40\% & 50\% \\
$\pi_{0.5}$~\citeyearpar{physicalintelligence2025pi05} & 10\% & 40\% & 20\% & 60\% \\
\midrule
\method{} (Task-only IDM) & 40\% & 75\% & 65\% & 75\% \\
\method{} (Full-play IDM) & \textbf{85\%} & \textbf{95\%} & \textbf{85\%} & \textbf{90\%} \\
\bottomrule
\end{tabular*}
\end{table}

\looseness=-1
\method{} achieves the highest mean success in all four conditions, and its
Task-only IDM also outperforms every external baseline.
Force-regulated sliding shows the largest gap. The Full-play IDM reaches
$85\%$ success. ImplicitRDP and FA-WAM reach $20\%$, and the other task-only
baselines reach $10$-$15\%$. A visual world-action model such as Cosmos Policy
can predict plausible motion, but the visual path alone does not specify how to
regulate contact force along it. Reactive feedback in RDP and ImplicitRDP, and
force-aware fusion in ForceVLA, expose the contact state but do not by themselves
learn a stronger goal-conditioned action response to force changes. Because these
methods rely on successful demonstrations, they may not observe enough variation
in such response behavior outside nominal contact conditions. \method{} combines
a task-conditioned wrench future with an IDM trained on task and play, giving
execution a contact reference and broader action-force experience.
The Sliding sequences in Figure~\ref{fig:real-robot-comparisons} illustrate
this contact demand: RDP and FA-WAM lose contact before completing the slide,
ForceVLA stalls near its initial position, and the shown \method{} rollout
maintains contact throughout the path and completes the task.

For box flipping, \method{} reaches $95\%$ success without disturbances and
$85\%$ with external pushes, compared with $60\%$ and $40\%$ for FA-WAM, the
strongest external comparator in both conditions. In Figure~\ref{fig:real-robot-comparisons},
\method{} retains tool-box contact through nominal and disturbed rotation,
while the baseline cases lose contact before completing the flip. An external
push can change the box-tool relationship, so repeating a nominal motion need
not preserve the contact needed for rotation. ForceWM supplies the anticipated
box motion and loading, while the IDM updates actions using measured wrench
feedback. This provides a mechanism for retaining the flipping objective while
adapting local execution, consistent with the higher success in both conditions.

Hanoi ring placement yields a different baseline ordering: Fast-WAM reaches
$65\%$, and Cosmos Policy, ForceVLA, and $\pi_{0.5}$ reach $60\%$, compared with
$90\%$ for \method{}. Their relatively strong placement results suggest that
visual alignment provides useful guidance here. Alignment alone, however,
does not determine whether the ring seats under contact. Conditioning the IDM
on the future seating geometry and accompanying wrench evolution gives local
force responses a specific placement outcome to pursue.
In the Hanoi comparison (Figure~\ref{fig:real-robot-comparisons}), \method{}
adjusts the ring pose and seats it on the pole. All three baseline examples
fail to complete placement.

\FloatBarrier
\Needspace{5\baselineskip}
\subsection{Q2: Does added task-agnostic play improve execution?}
\label{sec:play-scaling}

\begin{wrapfigure}[13]{R}{0.43\textwidth}
    \vspace{-10pt}
    \centering
    \includegraphics[width=0.94\linewidth,trim=0 5pt 0 5pt,clip]{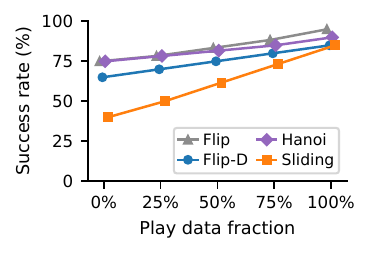}\par
    \setlength{\abovecaptionskip}{4pt}
    \caption{\textbf{Mean success with increasing play data.}
    Task demos, the ForceWM model, and IDM optimizer steps are fixed.}
    \label{fig:play-scaling}
\end{wrapfigure}

We train five IDMs with all task demonstrations and $0$, $50$, $100$, $150$,
or $200$ play episodes per task family, drawing the intermediate subsets
independently (Figure~\ref{fig:play-scaling}). ForceWM and the number of
optimizer updates are fixed across variants.

\looseness=-1
Force-regulated sliding shows the largest absolute gain, from $40\%$ to $85\%$
success in completing the path under the $5$-$15\,$N contact-force criterion.
This task repeatedly couples motion along a path with force regulation.
Successful demonstrations need not cover how to act when contact loads depart
from nominal conditions. Pressing, sliding, and release in play provide
additional action-wrench transitions for learning such responses. The IDM's
future condition associates these actions with their contact outcomes. The
large improvement is consistent with the value of this supervision for
sustained force regulation.

For box flipping, play raises success from $75\%$ to $95\%$ without disturbances
and from $65\%$ to $85\%$ with external pushes. Here, additional experience of
building and releasing contact can help the IDM adapt pushing actions as the
box rotates. Hanoi ring placement improves from $75\%$ to $90\%$: local
contact experience can likewise help select actions during seating, while the
task-trained future retains the required ring-pole geometry. These results
suggest that the benefit of play extends across different contact demands,
from continuous force regulation to rotation and placement.
With ForceWM held fixed, these gains reflect improved
execution learning from recorded play futures and actions, without additional
task demonstrations.

By contrast, the evaluated end-to-end models, ImplicitRDP and Cosmos Policy,
do not benefit from added play data: their shared policies learn directly from
play actions even when those actions need not advance the task. The added play
can therefore shift their action distributions away from successful task
behavior, including when Cosmos Policy also predicts futures. By decoupling
knowledge of task-compatible outcomes from knowledge of local action-contact
responses,
\method{} can learn from play in the IDM without training ForceWM on off-task
behavior.

\Needspace{5\baselineskip}
\subsection{Q3: Visualizing ForceWM predictions during closed-loop execution}
\label{sec:forcewm-quality}

\begingroup
\setlength{\intextsep}{2pt}
\begin{figure}[!htb]
    \centering
    \setlength{\abovecaptionskip}{2pt}
    \includegraphics[width=0.88\linewidth]{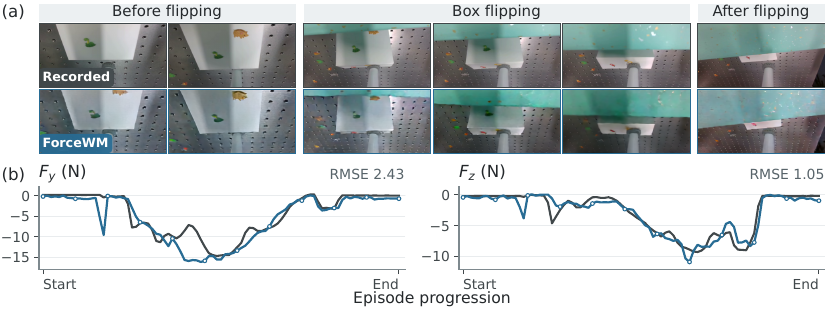}
    \caption{\textbf{ForceWM predictions during closed-loop box flipping.}
    \textbf{(a)}~Recorded wrist images (top) and decoded futures (bottom).
    The three central predictions share one conditioning observation.
    \textbf{(b)}~Recorded (gray) and predicted (blue) $F_y$ and $F_z$
    (all six wrench components in Appendix~\ref{app:forcewm-wrench}).}
    \label{fig:forcewm-quality}
\end{figure}
\endgroup

We visualize ForceWM predictions recorded during a closed-loop box-flipping
rollout on the real robot. Figure~\ref{fig:forcewm-quality} compares decoded visual futures and
predicted $F_y$ and $F_z$ with the wrist-camera observations and force measurements
recorded during execution.

Across the rollout, the visual futures capture the progression from approach
through box rotation to separation. During rotation, successive predicted
frames preserve the changing orientation of the box relative to the tool,
representing the intermediate geometry of the manipulation.
The force futures capture the contact loading associated with this motion:
load builds as the tool engages the box, remains substantial during rotation,
and subsides as the tool separates. Despite discrepancies in local visual
details and peak wrench values, the predictions preserve the main interaction
stages. This example illustrates how the generated future combines the
anticipated tool-object geometry with the accompanying contact loads to guide
the IDM.

\Needspace{5\baselineskip}
\subsection{Q4: Utility of visual-wrench future conditioning}
\label{sec:reference-executability}

\looseness=-1
Table~\ref{tab:reference-ablation} evaluates the utility of visual and wrench
future conditioning for online execution.  We independently train four shared multi-task
IDMs with visual$+$wrench futures, visual futures only, wrench futures only, or
no future conditioning.  All four retain the current wrist image and measured
wrench feedback and use the full task$+$play corpus.  The visual$+$wrench row
is the Full-play IDM of Table~\ref{tab:task-outcomes}.  Each future modality is
either retained or removed during both training and deployment.  The comparison with
no future conditioning tests the value of adding generated futures to current
observations. The single-modality variants assess the contributions of the
visual and wrench components.

\begingroup
\setlength{\intextsep}{6pt plus 2pt minus 2pt}
\begin{table}[!htb]
\setlength{\abovecaptionskip}{0pt}
\caption{\textbf{Online future-conditioning ablation: mean success rates.}
All IDMs use task$+$play data, with futures from one fixed ForceWM.
Rows 1-4 vary future modalities during training and deployment. The last
uses both without training-time future-conditioning augmentation.}
\label{tab:reference-ablation}
\centering
\footnotesize
\setlength{\tabcolsep}{4pt}
\renewcommand{\arraystretch}{1}
\begin{tabular*}{\linewidth}{@{\extracolsep{\fill}}lcccc@{}}
\toprule
\textbf{Future conditioning} & \textbf{Slide}$\,\uparrow$ & \textbf{Flip}$\,\uparrow$ & \textbf{Flip-D}$\,\uparrow$ & \textbf{Hanoi}$\,\uparrow$ \\
\midrule
Visual $+$ wrench & \textbf{85\%} & \textbf{95\%} & \textbf{85\%} & \textbf{90\%} \\
Visual only       & 15\% & 40\% & 30\% & 70\% \\
Wrench only       & 20\% & 30\% & 25\% & 30\% \\
No future conditioning & 15\% & 30\% & 35\% & 50\% \\
\midrule
Visual $+$ wrench, no augmentation & 65\% & 80\% & 70\% & 75\% \\
\bottomrule
\end{tabular*}
\end{table}
\endgroup

Joint visual-wrench conditioning achieves the highest mean success rate in
all four conditions.  Relative to visual-only conditioning, it improves success
by $55$, $55$, $20$, and $70$ percentage points on Flip, Flip-D, Hanoi, and
Sliding, respectively.  These gains show the value of including a wrench future
alongside the visual future.  Conversely, retaining the visual future alongside
the wrench future improves success by $60$-$65$ percentage points across
conditions.
Compared with no future conditioning, the joint condition improves success by
$40$-$70$ percentage points.  Neither individual future modality consistently
improves over no future conditioning. These results support complementary
guidance from visual and wrench futures under our training and deployment setup.

Finally, we ablate the future-conditioning augmentation introduced in
Section~\ref{sec:idm-learning} by retraining the joint visual-wrench IDM
without perturbations to its recorded future inputs (final row).
Both variants use ForceWM-generated futures at deployment.
Augmentation improves success by $15$ percentage points on Flip, Flip-D, and
Hanoi, and by $20$ points on Sliding. These gains show that regularizing
recorded future conditions improves execution with ForceWM-generated futures.

%% file: sections/figs/real_robot_comparisons.tex
\begin{figure}[!htb]
    \centering
    \includegraphics[width=0.95\linewidth]{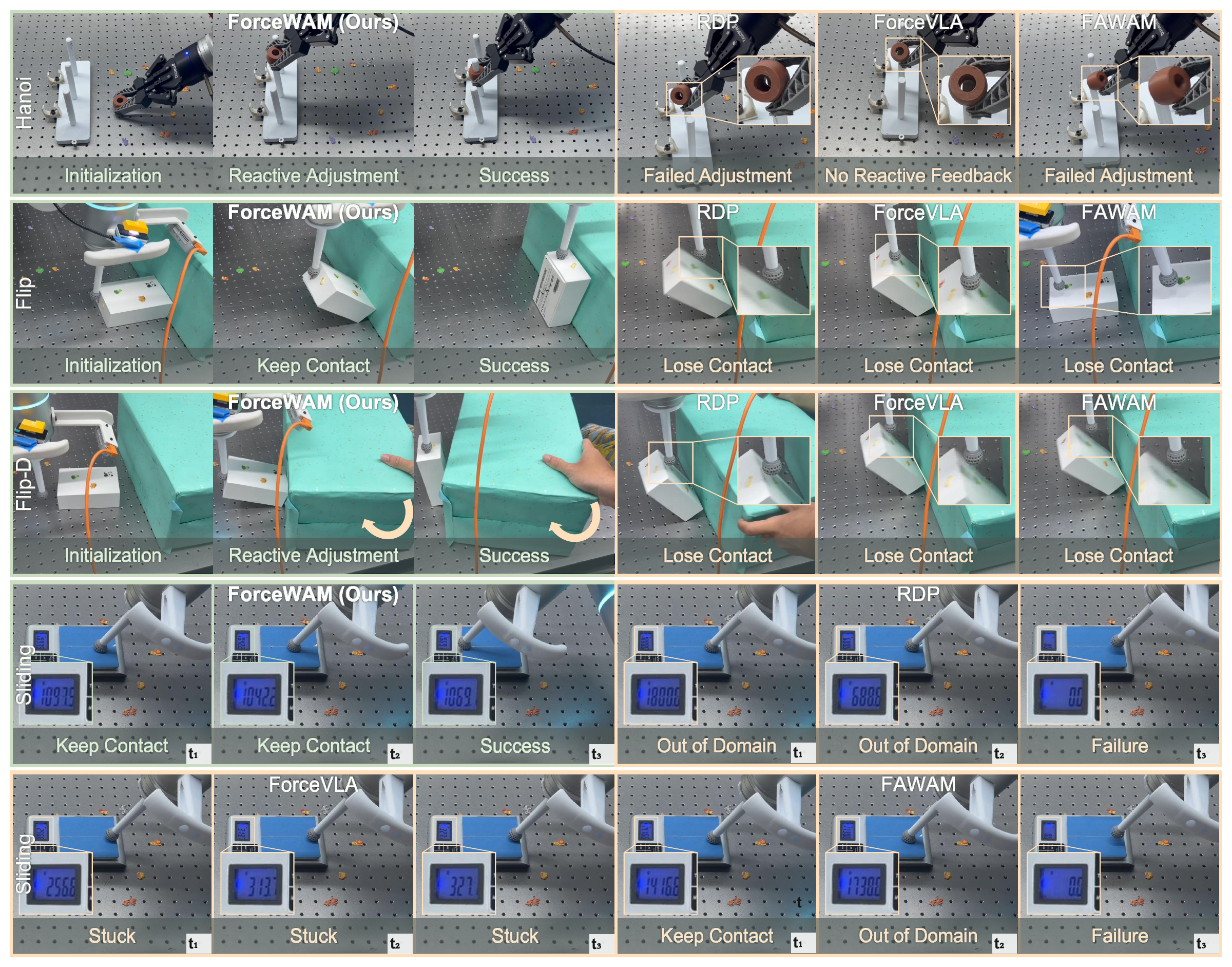}
    \caption{\textbf{Real-robot comparisons of \method{}, RDP, ForceVLA, and FA-WAM.}}
    \label{fig:real-robot-comparisons}
    \vspace{-8pt}
\end{figure}

%% file: sections/05_conclusion.tex
\section{Conclusion}
\label{sec:conclusion}

We presented \method{}, which decouples knowledge of task-compatible
interaction outcomes from knowledge of how local actions change contact. Task
demonstrations train a visual-wrench future model, while task and play data train an inverse
dynamics model that translates generated futures into actions with force
feedback. Real-robot experiments show higher task success than the evaluated
baselines and consistent gains from play. This suggests that using task and
play data to learn
complementary knowledge offers a promising path to scalable, stable
contact-rich manipulation.

%% file: sections/appendix.tex
\appendix

\section{Hardware setup and data collection}
\label{app:hardware}

All real-robot experiments use a single Flexiv Rizon 4S 7-DoF collaborative
arm equipped with joint torque sensors and a six-axis force/torque (F/T)
sensor at the end effector (Figure~\ref{fig:hardware-setup}).
The end-effector configuration depends on the task. Hanoi ring placement uses
a Robotiq 2F-85 parallel-jaw gripper. For box flipping (Flip and Flip-D) and
Sliding (Slide), a joystick and custom compliant fingertips are mounted at
the end effector.

We collect Hanoi data via virtual-reality (VR) teleoperation using a
Meta~Quest~3. For box flipping and Sliding, the joint
torque sensors support kinesthetic teaching during data collection, while the
end-effector F/T sensor records contact forces and torques. Across tasks,
this F/T sensor supplies the six-axis wrench observations used by ForceWM and
the IDM. A wrist-mounted RGB camera records visual observations. For Hanoi,
we also use a third-person camera.

\begingroup
\setlength{\intextsep}{6pt plus 2pt minus 2pt}
\begin{figure}[!htb]
    \centering
    \setlength{\abovecaptionskip}{2pt}
    \includegraphics[width=0.5\linewidth]{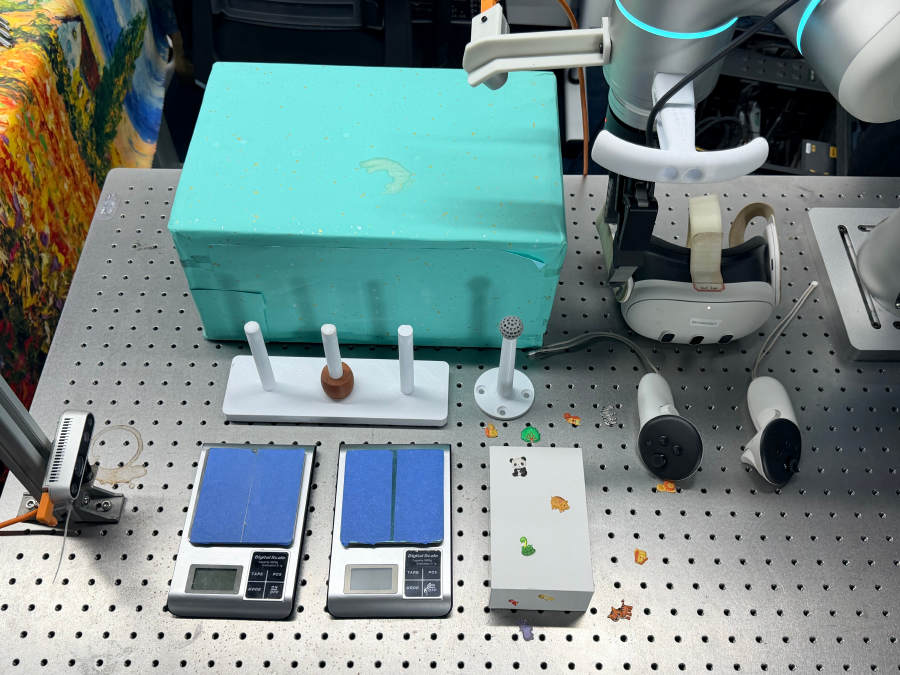}
    \caption{Hardware setup and task objects.}
    \label{fig:hardware-setup}
\end{figure}
\endgroup

\FloatBarrier
\section{Visual-wrench reference construction}
\label{app:representation}

Recorded training futures and ForceWM predictions use the same representation.
From the current image at time $t$, the visual target samples future wrist frames at
offsets $\{4,8,12,16\}$ and encodes them into one
$16\times56\times56$ video latent.  The wrench target contains the next 16
six-dimensional TCP wrench measurements.  Force and torque components are
normalized independently by a min-max map $n_F$ fitted on the ForceWM
training split.  The same frozen $n_F$ is used for recorded future trajectories
and online ForceWM predictions.  The measured causal wrench stream supplied to
the IDM uses a separate min-max map $n_I$ fitted on its training data. The
future-wrench stream remains in the $n_F$ coordinates and passes through its
IDM input normalizer unchanged.

ForceWM uses the temporal input convention of its pretrained video tokenizer.
Its 17-frame pixel sequence is organized as one blank frame, four copies of a
current-wrench placeholder, four copies of the current wrist image, four
future-wrench placeholders, and the four sampled future wrist frames.  The
video tokenizer temporally compresses this sequence into five
$16\times56\times56$ latent slots: blank, current wrench, current image, future
wrench, and future visual content.  The first three are clamped conditions.
Only the last two are noised and predicted.

A fixed adapter flattens the current $6$-D wrench or the future
$16\times6$ trajectory and periodically repeats it across the corresponding
video-latent-shaped slot.  Decoding averages the complete repeats for each
component.  This parameter-free packing lets the wrench and video slots share
the pretrained spatiotemporal backbone while keeping the two generated outputs
separate.  If a complete future modality crosses an episode boundary, its
generated slot is excluded from the loss rather than supervising on padded
values.

\section{Robot action representation}
\label{app:action-interface}

Each robot action comprises a Cartesian TCP-pose target and a gripper command.
In the training data, it is represented by the TCP pose and gripper state at
the next recorded timestep. Poses are expressed relative to the base TCP pose
at the start of the prediction window. Rotation is represented by roll, pitch,
and yaw.  At deployment, the predicted relative pose is composed with the
cached base TCP pose and converted to the absolute pose-gripper command accepted
by the robot controller.

The implementation inherits an auxiliary prediction target from ImplicitRDP
\citep{chen2026implicitrdp}.  In addition to the executable relative
pose-gripper action, it predicts a force-derived virtual pose and a scalar
stiffness.  This auxiliary block is trained jointly with the executable block,
but is discarded after inference: the robot receives only the decoded TCP pose
and gripper width. The auxiliary target regularizes representations learned
from the IDM's multimodal inputs. After
relative-pose conversion, the full diffusion target has 14 dimensions and an
18-step prediction horizon. Only the prefix aligned with the current slow
window is eligible for execution.

\section{Architecture and training details}
\label{app:training-details}

\paragraph{ForceWM.}
We initialize the spatiotemporal backbone from the Cosmos Predict 2B weights
\citep{nvidia2025cosmos} and train ForceWM directly on task demonstrations. No
intermediate task-specific pretraining is used.  The model takes a wrist-camera
image, a six-dimensional TCP wrench, and a task instruction (e.g., ``flip box''). Visual inputs
are encoded at $448\times448$.  The five-slot latent is denoised by the shared
Cosmos video diffusion backbone with frame-replacement conditioning.  A hybrid
EDM noise distribution draws 70\% of noise scales from the configured
log-normal distribution and 30\% uniformly from $[1,85]$. The sigma-weighted
MSE is evaluated only on the future-wrench and future-visual slots.  The task
instruction is supplied through its pretrained T5 text embedding.  We train
for 20,000 iterations on eight GPUs with batch size 8 per GPU, hold out the
final 30 episodes for validation, and save a checkpoint every 1,000
iterations.  The optimizer uses learning rate $3\times10^{-5}$.

\Needspace{9\baselineskip}
\paragraph{IDM.}
The IDM encodes the current $224\times224$ wrist image with a ResNet-34 whose
normalization layers are GroupNorm.  The resulting $7\times7$ feature map is
projected into 49 tokens of width 768, and the current 6-D wrench is linearly
projected into one additional token.  The $16\times56\times56$ future visual
latent is encoded by
$\operatorname{Conv}_{3\times3}(16,256)$-GroupNorm-SiLU, pooled to a
$2\times2$ map, and linearly projected into four 768-D tokens.  Each of the 16
future 6-D wrench samples is independently projected into one 768-D token.  A
one-layer, unidirectional GRU with hidden size 512 followed by a linear
projection produces a 768-D token for every measured wrench in the causal
prefix.

The action denoiser linearly embeds each 14-D noisy action and adds learned
temporal positions. The diffusion timestep uses a sinusoidal embedding.  A
learned projection and a two-layer Mish MLP encode the fixed condition memory.
A seven-layer pre-normalized Transformer decoder of width 768, four attention
heads, and $4\times$ feed-forward expansion predicts velocity for all 18 action
tokens during training.  Its action self-attention mask and its cross-attention mask
over GRU wrench tokens are both lower triangular.  A complementary mask hides
every future-wrench reference token whose window offset is less than or equal
to the queried action token.  Therefore action token $j$ can use noisy-action and
measured-wrench tokens $0{:}j$, the current image and wrench, reference-wrench
tokens at later window offsets, and all fixed future-visual tokens.  Processing the
full window in one forward pass is thus parallel teacher-forced training with
the same causal information boundary used by streaming inference, rather than
access to future measured feedback.  The action horizon is 18 and the model
observes one current image.  Deployment refreshes the slow reference after six
actions, so only tokens $0{:}5$ are eligible for execution. In particular, the two
tokens beyond the 16-step reference horizon are supervision-only.

For the play-scaling study, we sample training windows at random from the
pooled task and selected play trajectories.
The action model uses a 100-step squared-cosine diffusion schedule and velocity
prediction.
Training adds $\boldsymbol\epsilon+0.1\boldsymbol\epsilon'$ to the normalized
action while computing the velocity target from $\boldsymbol\epsilon$.
Inference uses five deterministic DDIM steps with $\eta=0$, leading timesteps
$[80,60,40,20,0]$, and clips the predicted normalized clean action to
$[-1,1]$ before denormalization.  Image augmentation consists of a center crop,
rotation in $[-5^\circ,5^\circ]$, and color jitter during training. Stochastic
image augmentation is disabled online.

To reduce the recorded-to-predicted future shift, IDM training applies
future-conditioning augmentation. It first selects the temporal lookup, then
drops the entire future condition with a fixed probability, followed by
independent modality dropout, additive
noise, and multiplicative scale jitter for the surviving visual and wrench
parts.  The exact flip-box values are summarized in
Table~\ref{tab:idm-reference-augmentation}.

\begingroup
\setlength{\intextsep}{6pt plus 2pt minus 2pt}
\begin{table}[!htb]
\setlength{\abovecaptionskip}{2pt}
\caption{Future-conditioning augmentation used for the flip-box IDM.}
\label{tab:idm-reference-augmentation}
\centering
\footnotesize
\setlength{\tabcolsep}{5pt}
\renewcommand{\arraystretch}{1}
\begin{tabular*}{\linewidth}{@{\extracolsep{\fill}}ll@{}}
\toprule
\textbf{Augmentation} & \textbf{Flip-box setting} \\
\midrule
Whole-condition dropout & $0.10$ \\
Visual / wrench dropout & $0.25$ / $0.50$ \\
Scale jitter & $[0.85,1.15]$ \\
Wrench noise std. & $0.075$ in normalized coordinates \\
Visual-latent noise std. & $0.075$ times latent standard deviation \\
Temporal offsets & $\{0,-1,-2,-3,-4,-5,-6,-9,-12\}$ \\
\bottomrule
\end{tabular*}
\end{table}
\endgroup

\FloatBarrier
\Needspace{5\baselineskip}
\section{Deployment details}
\label{app:deployment-details}

We deploy ForceWM and the IDM in a shared policy service.
ForceWM uses one denoising step and the IDM uses five. Measured separately
on the deployment system, their calls take approximately $100\,$ms and
$60\,$ms, respectively. The reference is refreshed every six actions.

Each refresh caches the generated visual-wrench reference, slow tokens,
initial IDM noise, and base TCP pose used for relative-to-absolute action
conversion. Between refreshes, the IDM deterministically re-denoises the
action prefix from cached noise using the
growing measured-wrench prefix, executing only its current action token.  The GRU
re-encodes the available wrench prefix at each call, and all reference-specific
caches reset at the next refresh.

At each reference refresh, ForceWM runs first, followed by the IDM.
The previous controller target remains active during these calls.
The first action after a refresh uses the
request-time wrench, and subsequent actions incorporate new measurements.
Reference-wrench visibility follows the action index within the window.

\FloatBarrier
\Needspace{5\baselineskip}
\section{Full six-axis wrench prediction comparison}
\label{app:forcewm-wrench}

Figure~\ref{fig:forcewm-wrench-full} extends the $F_y$ and $F_z$ visualization
in Figure~\ref{fig:forcewm-quality} to all six wrench components from the same
closed-loop box-flipping rollout.

\begingroup
\setlength{\intextsep}{6pt plus 2pt minus 2pt}
\begin{figure}[H]
    \centering
    \setlength{\abovecaptionskip}{2pt}
    \includegraphics[width=0.9\linewidth]{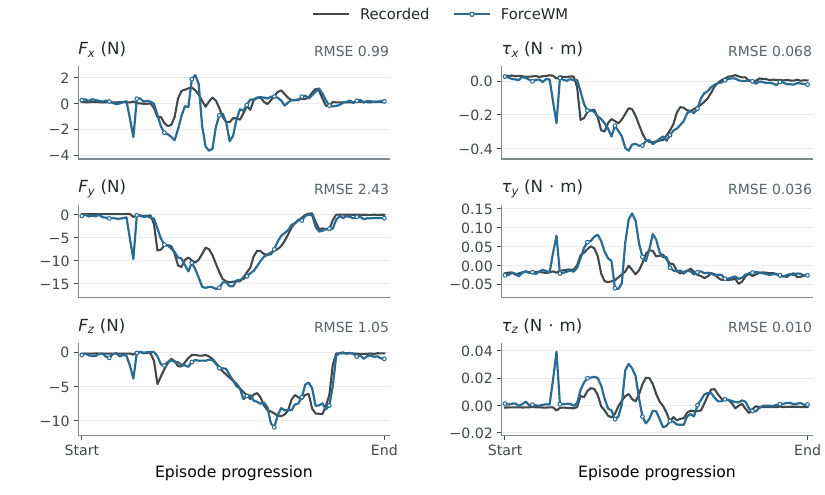}
    \caption{Full six-axis comparison for the rollout in
    Figure~\ref{fig:forcewm-quality}. Recorded (gray) and predicted (blue)
    forces are shown on the left and torques on the right.}
    \label{fig:forcewm-wrench-full}
\end{figure}
\endgroup

Beyond the force profiles in the main paper, the predicted $\tau_x$
also captures the overall torque buildup and release during box flipping.
Together with the visual futures in Figure~\ref{fig:forcewm-quality}, these
predictions describe the evolving tool-object geometry and the accompanying
interaction loads, providing complementary references for the IDM.

%% file: forcewam.bib
@misc{ye2026dreamzero,
  title         = {World Action Models are Zero-shot Policies},
  author        = {Ye, Seonghyeon and Ge, Yunhao and Zheng, Kaiyuan and Gao, Shenyuan and Yu, Sihyun and Kurian, George and Indupuru, Suneel and Tan, You Liang and Zhu, Chuning and Xiang, Jiannan and Malik, Ayaan and Lee, Kyungmin and Liang, William and Ranawaka, Nadun and Gu, Jiasheng and Xu, Yinzhen and Wang, Guanzhi and Hu, Fengyuan and Narayan, Avnish and Bjorck, Johan and Wang, Jing and Kim, Gwanghyun and Niu, Dantong and Zheng, Ruijie and Xie, Yuqi and Wu, Jimmy and Wang, Qi and Julian, Ryan and Xu, Danfei and Du, Yilun and Chebotar, Yevgen and Reed, Scott and Kautz, Jan and Zhu, Yuke and Fan, Linxi Jim and Jang, Joel},
  year          = {2026},
  howpublished  = {arXiv preprint arXiv:2602.15922},
  eprint        = {2602.15922},
  archiveprefix = {arXiv},
  primaryclass  = {cs.RO}
}

@inproceedings{bi2026motus,
  title     = {Motus: A Unified Latent Action World Model},
  author    = {Bi, Hongzhe and Tan, Hengkai and Xie, Shenghao and Wang, Zeyuan and Huang, Shuhe and Liu, Haitian and Zhao, Ruowen and Feng, Yao and Xiang, Chendong and Rong, Yinze and Zhao, Hongyan and Liu, Hanyu and Su, Zhizhong and Ma, Lei and Su, Hang and Zhu, Jun},
  booktitle = {Proceedings of the IEEE/CVF Conference on Computer Vision and Pattern Recognition},
  year      = {2026},
  pages     = {35101--35113}
}

@article{hogan1985impedance,
  title   = {Impedance Control: An Approach to Manipulation: Part {I}---Theory},
  author  = {Hogan, Neville},
  journal = {Journal of Dynamic Systems, Measurement, and Control},
  volume  = {107},
  number  = {1},
  pages   = {1--7},
  year    = {1985},
  doi     = {10.1115/1.3140702}
}

@inproceedings{chi2023diffusionpolicy,
  title     = {Diffusion Policy: Visuomotor Policy Learning via Action Diffusion},
  author    = {Chi, Cheng and Feng, Siyuan and Du, Yilun and Xu, Zhenjia and Cousineau, Eric and Burchfiel, Benjamin and Song, Shuran},
  booktitle = {Robotics: Science and Systems},
  year      = {2023},
  doi       = {10.15607/RSS.2023.XIX.026}
}

@inproceedings{karras2022elucidating,
  title     = {Elucidating the Design Space of Diffusion-Based Generative Models},
  author    = {Karras, Tero and Aittala, Miika and Aila, Timo and Laine, Samuli},
  booktitle = {Advances in Neural Information Processing Systems},
  volume    = {35},
  pages     = {26565--26577},
  year      = {2022}
}

@inproceedings{liu2025forcemimic,
  title         = {{ForceMimic}: Force-Centric Imitation Learning with Force-Motion Capture System for Contact-Rich Manipulation},
  author        = {Liu, Wenhai and Wang, Junbo and Wang, Yiming and Wang, Weiming and Lu, Cewu},
  booktitle     = {IEEE International Conference on Robotics and Automation},
  year          = {2025},
  eprint        = {2410.07554},
  archiveprefix = {arXiv},
  primaryclass  = {cs.RO}
}

@inproceedings{xue2025rdp,
  title         = {Reactive Diffusion Policy: Slow-Fast Visual-Tactile Policy Learning for Contact-Rich Manipulation},
  author        = {Xue, Han and Ren, Jieji and Chen, Wendi and Zhang, Gu and Fang, Yuan and Gu, Guoying and Xu, Huazhe and Lu, Cewu},
  booktitle     = {Robotics: Science and Systems},
  year          = {2025},
  doi           = {10.15607/RSS.2025.XXI.052},
  eprint        = {2503.02881},
  archiveprefix = {arXiv},
  primaryclass  = {cs.RO}
}

@inproceedings{yu2025forcevla,
  title         = {{ForceVLA}: Enhancing {VLA} Models with a Force-aware {MoE} for Contact-rich Manipulation},
  author        = {Yu, Jiawen and Liu, Hairuo and Yu, Qiaojun and Ren, Jieji and Hao, Ce and Ding, Haitong and Huang, Guangyu and Huang, Guofan and Song, Yan and Cai, Panpan and Zhang, Wenqiang and Lu, Cewu},
  booktitle     = {Advances in Neural Information Processing Systems},
  volume        = {38},
  pages         = {93409--93439},
  year          = {2025},
  doi           = {10.52202/085713-3124},
  eprint        = {2505.22159},
  archiveprefix = {arXiv},
  primaryclass  = {cs.RO}
}

@inproceedings{hou2024acp,
  title     = {Adaptive Compliance Policy: Learning Approximate Compliance for Diffusion Guided Control},
  author    = {Hou, Yifan and Liu, Zeyi and Chi, Cheng and Cousineau, Eric and Kuppuswamy, Naveen and Feng, Siyuan and Burchfiel, Benjamin and Song, Shuran},
  booktitle = {IEEE International Conference on Robotics and Automation},
  pages     = {4829--4836},
  year      = {2025},
  doi       = {10.1109/ICRA55743.2025.11128452}
}

@inproceedings{fang2026forcepolicy,
  title     = {{Force Policy}: Learning Hybrid Force-Position Control Policy under Interaction Frame for Contact-Rich Manipulation},
  author    = {Fang, Hongjie and Tang, Shirun and Mei, Mingyu and Qin, Haoxiang and He, Zihao and Chen, Jingjing and Feng, Ying and Wang, Chenxi and Liu, Wanxi and He, Zaixing and Lu, Cewu and Wang, Shiquan},
  booktitle = {Robotics: Science and Systems},
  year      = {2026},
  doi       = {10.15607/RSS.2026.XXII.128}
}

@inproceedings{li2026forcevla2,
  title     = {{ForceVLA2}: Unleashing Hybrid Force-Position Control with Force Awareness for Contact-Rich Manipulation},
  author    = {Li, Yang and {Zhaxizhuoma} and Jiang, Hongru and Xia, Junjie and Zhang, Hongquan and Du, Jinda and Zhou, Yunsong and Zeng, Jia and Hao, Ce and Ren, Jieji and Yu, Qiaojun and Lu, Cewu and Qiao, Yu and Pang, Jiangmiao},
  booktitle = {Proceedings of the IEEE/CVF Conference on Computer Vision and Pattern Recognition},
  pages     = {8911--8920},
  year      = {2026}
}

@article{chen2026implicitrdp,
  title   = {{ImplicitRDP}: An End-to-End Visual-Force Diffusion Policy with Structural Slow-Fast Learning},
  author  = {Chen, Wendi and Xue, Han and Wang, Yi and Zhou, Fangyuan and Lv, Jun and Jin, Yang and Tang, Shirun and Wen, Chuan and Lu, Cewu},
  journal = {IEEE Robotics and Automation Letters},
  volume  = {11},
  number  = {8},
  pages   = {10010--10017},
  year    = {2026},
  doi     = {10.1109/LRA.2026.3710031}
}

@misc{zhuo2026fardp,
  title         = {{FA-RDP}: A Frequency-Adaptive Reactive Diffusion Policy for Contact-Rich Manipulation},
  author        = {Zhuo, Lifeng and Chen, Wendi and Xue, Han and Tang, Shirun and Lv, Jun and Lu, Cewu and Wen, Chuan},
  year          = {2026},
  howpublished  = {arXiv preprint arXiv:2607.28596},
  eprint        = {2607.28596},
  archiveprefix = {arXiv},
  primaryclass  = {cs.RO}
}

@inproceedings{wang2026lift,
  title     = {Never Too Late for Force: Accelerating {VLA} Post-Training with Reactive Force Injection},
  author    = {Wang, Yi and Chen, Wendi and Wen, Zimo and Xue, Han and Li, Xueqi and Yu, Wenye and Chen, Zhijie and Yang, Hao and Lv, Jun and Wen, Chuan and Lu, Cewu},
  booktitle = {Proceedings of the Conference on Robot Learning},
  year      = {2026},
  note      = {Accepted}
}

@inproceedings{li2026lingbotva,
  title     = {Causal World Modeling for Robot Control},
  author    = {Li, Lin and Zhang, Qihang and Luo, Yiming and Yang, Shuai and Wang, Ruilin and Zhang, Luyao and Yu, Mingrui and Gao, Zelin and Xue, Nan and Zhou, Boyu and Zhu, Xing and Ding, Mingyu and Shen, Yujun and Xu, Yinghao},
  booktitle = {Robotics: Science and Systems},
  year      = {2026},
  doi       = {10.15607/RSS.2026.XXII.016}
}

@inproceedings{kim2026cosmospolicy,
  title     = {Cosmos Policy: Fine-Tuning Video Models for Visuomotor Control and Planning},
  author    = {Kim, Moo Jin and Gao, Yihuai and Lin, Tsung-Yi and Lin, Yen-Chen and Ge, Yunhao and Lam, Grace and Liang, Percy and Song, Shuran and Liu, Ming-Yu and Finn, Chelsea and Gu, Jinwei},
  booktitle = {International Conference on Learning Representations},
  year      = {2026}
}

@misc{nvidia2025cosmos,
  title         = {Cosmos World Foundation Model Platform for Physical {AI}},
  author        = {{NVIDIA} and Agarwal, Niket and Ali, Arslan and Bala, Maciej and Balaji, Yogesh and others},
  year          = {2025},
  howpublished  = {arXiv preprint arXiv:2501.03575},
  eprint        = {2501.03575},
  archiveprefix = {arXiv},
  primaryclass  = {cs.CV}
}

@misc{higuera2026vtwm,
  title         = {Visuo-Tactile World Models},
  author        = {Higuera, Carolina and Arnaud, Sergio and Boots, Byron and Mukadam, Mustafa and Hogan, Francois Robert and Meier, Franziska},
  year          = {2026},
  howpublished  = {arXiv preprint arXiv:2602.06001},
  eprint        = {2602.06001},
  archiveprefix = {arXiv},
  primaryclass  = {cs.RO}
}

@misc{zheng2026omnivta,
  title         = {{OmniVTA}: Visuo-Tactile World Modeling for Contact-Rich Robotic Manipulation},
  author        = {Zheng, Yuhang and Gu, Songen and Zheng, Yupeng and Li, Weize and Zang, Yujie and Tian, Shuai and Li, Xiang and Hao, Ce and Gao, Chen and Liu, Si and Li, Haoran and Chen, Yilun and Yan, Shuicheng and Ding, Wenchao},
  year          = {2026},
  howpublished  = {arXiv preprint arXiv:2603.19201},
  eprint        = {2603.19201},
  archiveprefix = {arXiv},
  primaryclass  = {cs.RO}
}

@misc{he2026fawam,
  title         = {{FAWAM}: Force-Aware World Action Models for Closed-Loop Contact-Rich Manipulation},
  author        = {He, Haotian and Yan, Zeyu and Liu, Qipeng and Guo, Ning and Lian, Wenzhao},
  year          = {2026},
  howpublished  = {arXiv preprint arXiv:2606.08555},
  eprint        = {2606.08555},
  archiveprefix = {arXiv},
  primaryclass  = {cs.RO}
}

@misc{yuan2026fastwam,
  title         = {{Fast-WAM}: Do World Action Models Need Test-time Future Imagination?},
  author        = {Yuan, Tianyuan and Dong, Zibin and Liu, Yicheng and Zhao, Hang},
  year          = {2026},
  howpublished  = {arXiv preprint arXiv:2603.16666},
  eprint        = {2603.16666},
  archiveprefix = {arXiv},
  primaryclass  = {cs.CV}
}

@inproceedings{physicalintelligence2025pi05,
  title     = {{$\pi_{0.5}$}: a Vision-Language-Action Model with Open-World Generalization},
  author    = {Black, Kevin and Brown, Noah and Darpinian, James and Dhabalia, Karan and Driess, Danny and Esmail, Adnan and Equi, Michael and Finn, Chelsea and Fusai, Niccolo and Galliker, Manuel Y. and Ghosh, Dibya and Groom, Lachy and Hausman, Karol and Ichter, Brian and Jakubczak, Szymon and Jones, Tim and Ke, Liyiming and LeBlanc, Devin and Levine, Sergey and Li-Bell, Adrian and Mothukuri, Mohith and Nair, Suraj and Pertsch, Karl and Ren, Allen Z. and Shi, Lucy Xiaoyang and Smith, Laura and Springenberg, Jost Tobias and Stachowicz, Kyle and Tanner, James and Vuong, Quan and Walke, Homer and Walling, Anna and Wang, Haohuan and Yu, Lili and Zhilinsky, Ury},
  booktitle = {Proceedings of the Conference on Robot Learning},
  volume    = {305},
  series    = {Proceedings of Machine Learning Research},
  pages     = {17--40},
  year      = {2025},
  publisher = {PMLR}
}

@misc{zhou2026touchworld,
  title         = {{TouchWorld}: A Predictive and Reactive Tactile Foundation Model for Dexterous Manipulation},
  author        = {Zhou, Jianyi and Hong, Feiyang and Li, Yunhao and Zhao, Yicheng and Cen, Yongjue and Liu, Zirui and Huang, Jiakang and Chen, Zirui and Zhang, Ruiyang and Zhu, Weizhuo and Song, Xuhua and Yang, Shuo},
  year          = {2026},
  howpublished  = {arXiv preprint arXiv:2607.07287},
  eprint        = {2607.07287},
  archiveprefix = {arXiv},
  primaryclass  = {cs.RO}
}

@inproceedings{du2023unipi,
  title     = {Learning Universal Policies via Text-Guided Video Generation},
  author    = {Du, Yilun and Yang, Mengjiao and Dai, Bo and Dai, Hanjun and Nachum, Ofir and Tenenbaum, Joshua B. and Schuurmans, Dale and Abbeel, Pieter},
  booktitle = {Advances in Neural Information Processing Systems},
  year      = {2023}
}

@misc{li2026vera,
  title         = {Turning Video Models into Generalist Robot Policies},
  author        = {Li, Sizhe Lester and Kim, Evan and Bai, Xingjian and Zhao, Tong and Pang, Tao and Simchowitz, Max and Sitzmann, Vincent},
  year          = {2026},
  howpublished  = {arXiv preprint arXiv:2605.27817},
  eprint        = {2605.27817},
  archiveprefix = {arXiv},
  primaryclass  = {cs.RO}
}

@misc{cai2026ahawam,
  title         = {{AHA-WAM}: Asynchronous Horizon-Adaptive World-Action Modeling with Observation-Guided Context Routing},
  author        = {Cai, Jisong and Ling, Long and Chu, Shiwei and Liu, Zhongshan and Kang, Jiayue and Liang, Zhixuan and Xu, Wenjie and Mao, Yinan and Zhang, Weinan and Yang, Xiaokang and Ying, Ru and Zheng, Ran and Mu, Yao},
  year          = {2026},
  howpublished  = {arXiv preprint arXiv:2606.09811},
  eprint        = {2606.09811},
  archiveprefix = {arXiv},
  primaryclass  = {cs.RO}
}

@misc{neoteai2026n0twam,
  title         = {{$N_0$-{TWAM}}: Scaling Tactile-Native World-Action Model for Contact-Rich Manipulation},
  author        = {{NeoteAI Team} and {Fudan TEAI Team}},
  year          = {2026},
  howpublished  = {arXiv preprint arXiv:2607.23783},
  eprint        = {2607.23783},
  archiveprefix = {arXiv},
  primaryclass  = {cs.RO}
}

@misc{bi2026motus2,
  title         = {{Motus2}: A Self-Evolving General World Model for Dexterous Manipulation},
  author        = {Bi, Hongzhe and Zhou, Zihao and Tang, Yihang and Pang, Jingrui and Huang, Shuhe and Liu, Haitian and Wang, Runqing and Huang, Shuai and Wang, Yichen and Cheng, Yiming and Zhao, Ruowen and Li, Zhenghua and Tan, Hengkai and Liu, Xiaolong and Wan, Jinhui and Liu, Jiabao and Zhao, Min and Bao, Fan and Zhu, Jun},
  year          = {2026},
  howpublished  = {arXiv preprint arXiv:2608.30237},
  eprint        = {2608.30237},
  archiveprefix = {arXiv},
  primaryclass  = {cs.RO}
}

@misc{ma2026tacpac,
  title         = {{TacPAC}: Tactile Prediction and Real-Time Action Correction in World-Action Models for Contact-Rich Manipulation},
  author        = {Ma, Zipei and Wei, Xiaofei and Jiang, Junzhe and Lu, Shunlin and Zhang, Li},
  year          = {2026},
  howpublished  = {arXiv preprint arXiv:2609.05266},
  eprint        = {2609.05266},
  archiveprefix = {arXiv},
  primaryclass  = {cs.RO}
}

@misc{peng2026fact,
  title         = {{FACT}: Failure-Aware Causal Training for World-Action Models},
  author        = {Peng, Quanquan and Liang, Yutong and Yan, Rui and Hansen, Nicklas and Wang, Xiaolong},
  year          = {2026},
  howpublished  = {arXiv preprint arXiv:2608.10232},
  eprint        = {2608.10232},
  archiveprefix = {arXiv},
  primaryclass  = {cs.RO}
}

@misc{agibot2026geact2,
  title         = {{GE-Act 2.0}: Pretraining and Scaling a World-Action Model for Robotic Manipulation},
  author        = {{AgiBot Research Team} and Liu, Renhang and Zhao, Wenzhi and Yang, Zhuo and Chen, Liliang and Zhou, Pengfei and Chen, Shengcong and Ren, Guanghui and Peng, Youlun and Jin, Rongjun and Wang, Nan and Wang, Sukai and He, Xindong and Feng, Jinyuan and Xiong, Ziyu and Zhong, Linqing and Wei, Yifei and Han, Feng and Zhang, Long and Huang, Da and Zhao, Nanshu and Yin, Chenghao and Wu, Mo and Yan, Zhaodong and Hu, Kongtao and Yan, Yuxiang and Niyazi, Aogelijiang and Fang, Yu and Zeng, Jia and Meng, Lizhu and Lv, Daizhen and Cao, Haoyu and Hou, Zhiwen and Ye, Lianjin and Niu, Yuehan and Cai, Zhikai and Hu, Xuan and Min, Hui and Cai, Xiongfeng and Liao, Yue and Wu, Jing and Poria, Soujanya and Li, Ye and Zhou, Sanping and Yao, Maoqing},
  year          = {2026},
  howpublished  = {arXiv preprint arXiv:2609.05588},
  eprint        = {2609.05588},
  archiveprefix = {arXiv},
  primaryclass  = {cs.RO}
}

@inproceedings{lee2019makingsense,
  title     = {Making Sense of Vision and Touch: Self-Supervised Learning of Multimodal Representations for Contact-Rich Tasks},
  author    = {Lee, Michelle A. and Zhu, Yuke and Srinivasan, Krishnan and Shah, Parth and Savarese, Silvio and Fei-Fei, Li and Garg, Animesh and Bohg, Jeannette},
  booktitle = {IEEE International Conference on Robotics and Automation},
  year      = {2019}
}

@article{suresh2024neuralfeels,
  title   = {{NeuralFeels} with Neural Fields: Visuotactile Perception for In-Hand Manipulation},
  author  = {Suresh, Sudharshan and Qi, Haozhi and Wu, Tingfan and Fan, Taosha and Pineda, Luis and Lambeta, Mike and Malik, Jitendra and Kalakrishnan, Mrinal and Calandra, Roberto and Kaess, Michael and Ortiz, Joseph and Mukadam, Mustafa},
  journal = {Science Robotics},
  volume  = {9},
  number  = {96},
  pages   = {eadl0628},
  year    = {2024},
  doi     = {10.1126/scirobotics.adl0628}
}

@inproceedings{ai2024robopack,
  title     = {{RoboPack}: Learning Tactile-Informed Dynamics Models for Dense Packing},
  author    = {Ai, Bo and Tian, Stephen and Shi, Haochen and Wang, Yixuan and Tan, Cheston and Li, Yunzhu and Wu, Jiajun},
  booktitle = {Robotics: Science and Systems},
  year      = {2024}
}

@article{calandra2018feeling,
  title   = {More Than a Feeling: Learning to Grasp and Regrasp Using Vision and Touch},
  author  = {Calandra, Roberto and Owens, Andrew and Jayaraman, Dinesh and Lin, Justin and Yuan, Wenzhen and Malik, Jitendra and Adelson, Edward H. and Levine, Sergey},
  journal = {IEEE Robotics and Automation Letters},
  volume  = {3},
  number  = {4},
  pages   = {3300--3307},
  year    = {2018},
  doi     = {10.1109/LRA.2018.2852779}
}

@article{xu2025vilp,
  title   = {{VILP}: Imitation Learning with Latent Video Planning},
  author  = {Xu, Zhengtong and Qiu, Qiang and She, Yu},
  journal = {IEEE Robotics and Automation Letters},
  volume  = {10},
  number  = {4},
  pages   = {3350--3357},
  year    = {2025},
  doi     = {10.1109/LRA.2025.3542317}
}

@inproceedings{madan2026tactic,
  title     = {{TACTIC}: Tactile and Vision Conditioned Contact-Centric Control for Whole-Arm Manipulation},
  author    = {Madan, Rishabh and Xie, Angchen and Saak, Samantha and Blanco, Andres and Lee, Dohyeok and Brown, Sarah Grace and Yan, Yunting and Zolotas, Mark and Barreiros, Jose and Bhattacharjee, Tapomayukh},
  booktitle = {Robotics: Science and Systems},
  year      = {2026},
  doi       = {10.15607/RSS.2026.XXII.060}
}
